\documentclass[10pt,twocolumn,letterpaper]{article}

\usepackage{cvpr}
\usepackage{times}
\usepackage{epsfig}
\usepackage{graphicx}
\usepackage{amsmath}
\usepackage{amssymb}
\usepackage{cvpr}
\usepackage{times}
\usepackage{epsfig}
\usepackage{graphicx}
\usepackage{amsmath,bm}
\usepackage{amssymb}
\usepackage{booktabs}
\usepackage{bbm}
\usepackage{multirow}
\usepackage{tabularx}
\usepackage{authblk}

\usepackage[pagebackref=true,breaklinks=true,letterpaper=true,colorlinks,bookmarks=false]{hyperref}

\cvprfinalcopy 

\def\cvprPaperID{****} 

\begin{document}

\title{V2F-Net: Explicit Decomposition of Occluded Pedestrian Detection}
\makeatletter
\renewcommand*{\@fnsymbol}[1]{\ensuremath{\ifcase#1\or *\or \dagger\or \ddagger\or
    \mathsection\or \mathparagraph\or \|\or **\or \dagger\dagger
    \or \ddagger\ddagger \else\@ctrerr\fi}}
\makeatother
\newcommand*\samethanks[1][\value{footnote}]{\footnotemark[#1]}

\author[1]{Mingyang Shang\thanks{The first two authors contribute equally to this work.}}
\author[1,2]{Dawei Xiang\samethanks\thanks{This work was done when Dawei Xiang was an intern at MEGVII Tech.}}
\author[1]{Zhicheng Wang}
\author[1]{Erjin Zhou}
\affil[1]{MEGVII Technology} 
\affil[2]{Texas A\&M University}
\affil[ ]{
\tt\small shangmingyang@megvii.com \quad
\tt\small xiangdw@tamu.edu \quad
\tt\small wangzhicheng@megvii.com \quad
\tt\small zej@megvii.com
}
\setlength{\affilsep}{0em}
\renewcommand\Authsep{\quad}
\renewcommand\Authand{\quad}
\renewcommand\Authands{\quad}

\maketitle

\begin{abstract}

   Occlusion is very challenging in pedestrian detection. In this paper, we propose a simple yet effective method named V2F-Net, which explicitly decomposes occluded pedestrian detection into visible region detection and full body estimation.
  V2F-Net consists of two sub-networks: Visible region Detection Network (VDN) and Full body Estimation Network (FEN).
  VDN tries to localize visible regions and FEN estimates full-body box on the basis of the visible box.
  Moreover, to further improve the estimation of full body, we propose a novel Embedding-based Part-aware Module (EPM).
  By supervising the visibility for each part, the network is encouraged to extract features with essential part information. We experimentally show the effectiveness of V2F-Net by conducting several experiments on two challenging datasets. V2F-Net achieves 5.85\% AP gains on CrowdHuman and 2.24\% $\text{MR}^{-2}$ improvements on CityPersons compared to FPN baseline. Besides, the consistent gain on both one-stage and two-stage detector validates the generalizability of our method.

\end{abstract}

\section{Introduction}


Pedestrian detection is an important task in computer vision. It is a  fundamental component of many applications, such as video surveillance, autonomous driving, etc. Benefiting from the development of deep learning in recent years, especially the proposal of Convolutional Neural Network~(CNN) based general objects detection methods~\cite{girshick2014rich,ren2015faster,he2017mask,yolov3,liu2016ssd,dai2016rfcn,lin2017feature,lin2017focal,cai2019cascadercnn}, many works~\cite{optimizedpedestrian,tian2015deep,stewart2016end,zhou2017multilabelpart,discriminativeoccluding,Tang2014,chi2019pedhunter} have been done to adapt them to pedestrian detection, leading to great progress in this field.

\begin{figure}[!t]
\begin{center}
 \includegraphics[width=1.\linewidth]{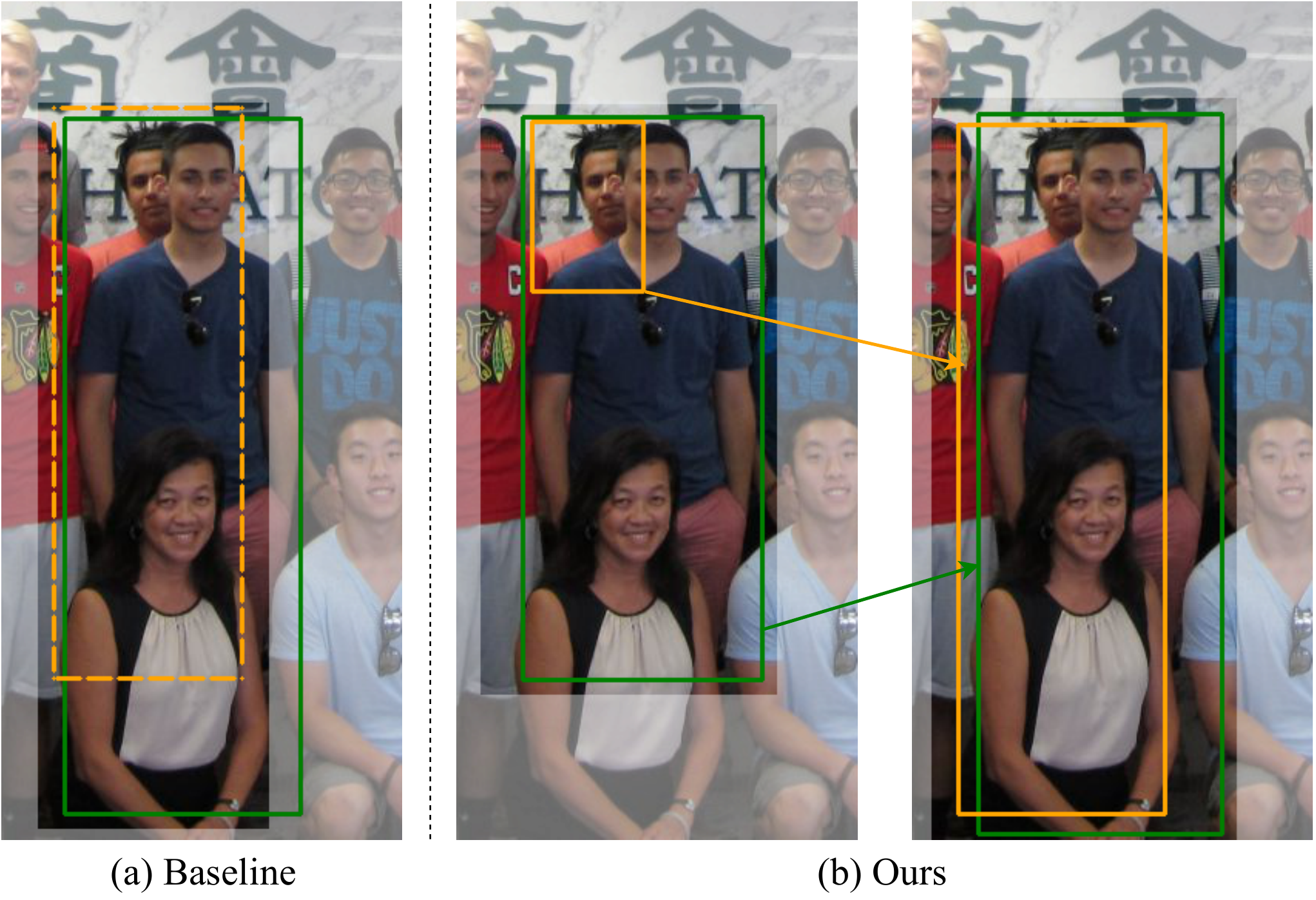}
\end{center}
   \caption{Illustration of V2F-Net. (a) Results of Faster-RCNN~\cite{ren2015faster} with FPN~\cite{lin2017feature} baseline. (b) Results of our method. The solid boxes indicate detected boxes and the dashed boxes indicate false suppression by NMS. Boxes belong to the same pedestrian are drawn in the same color. Arrow in Fig.~\ref{fig:illustration} (b) represents full body estimation from visible region. Directly detecting full body pedestrian tends to cause inaccurate regression; meanwhile, using full boxes to perform NMS is responsible for false suppression. Our method predicts visible box and full-body box of pedestrian sequentially, and we do calculation of IoU in NMS on visible boxes instead of full boxes. Therefore, both the two full boxes are precise and can be kept even IoU between them is high. Best viewed in color.}
\label{fig:illustration}
\end{figure}

However, occlusion remains a big challenge in detecting pedestrian. Many efforts have been made to handle this problem. Some of prior works ~\cite{wang2018repulsion,discriminativeoccluding,chu2020crowddet} only use the full-body box of pedestrian to train their networks, which implies an assumption more or less: the annotated full-body box is visible completely~\cite{pang2019mask}. However, this assumption may not be true in the occlusion cases. Therefore, some other researchers try to leverage visible box to assist the full-body pedestrian detection~\cite{pang2019mask,zhang2018Occlusionaware,Zhou_2018_bibox,zhang2018occludedattention,pang2019mask,huang2020R2nms}, and have achieved inspiring improvements.  


Intuitively, the process of pedestrian detection will go through two stages for human: \textbf{detect} visible region by ``seeing'' first, then \textbf{estimate} full body based on the structure and proportion of the human body. The visible region is critical in this process, as it is a strong evidence for identifying person and discriminating between two different pedestrians. However, prior works adopt different pipelines from the process of human vision.~\cite{wang2018repulsion, chu2020crowddet} predict only full body box directly, while~\cite{Zhou_2018_bibox, huang2020R2nms} output visible box and full box parallelly. Both of them consider the full-body pedestrian detection as a single problem. Obviously this would increase the difficulty of network learning.~\cite{prnet} takes visible box as an auxiliary priori to initialize the full-body anchors. It still fails to realize the
key role of visible box, thus will suffer from false suppression by NMS severely and be limited to datasets with fixed ratios of full body annotation.

Inspired by above observation, we propose a new solution from the perspective of pipeline: V2F-Net. We decompose occluded pedestrian detection into two sub-problems explicitly: visible region detection and full body estimation from visible region. Each sub-problem corresponds to a sub-network of V2F-Net, called Visible region Detection Network (VDN) and Full body Estimation Network (FEN) respectively. V2F-Net has a straightforward pipeline and can be trained end-to-end. An input image is first processed by VDN to detect visible boxes of all pedestrians, after \textit{Non-maximum Suppression} (NMS) these kept boxes are fed into FEN to estimate final full-body box for each pedestrian. Fig.~\ref{fig:illustration} shows an illustration of V2F-Net.

One may doubt that detecting visible box is harder than full-body box, as reported in \cite{shao2018crowdhuman}. Based on our observations, we found that the higher $\text{MR}^{-2}$ of visible box detection is mostly because of relatively lower quality of detected visible boxes. This can be partly validated from Table~\ref{tbl:crowdhuman_iou}. If we loose the matching \textit{Intersection over Union} (IoU) threshold between detected box and ground truth, both AP and $\text{MR}^{-2}$ of visible boxes can outperform the results of full-body boxes, which indicates the location of detected visible boxes are not precise as full-body boxes. We suspect there are two main reasons for it. Compared to full boxes, visible boxes have a \textit{wider range of aspect ratio} and are usually \textit{smaller}. The former causes regression offsets with larger variance, and the latter 
leads to higher susceptibility to minor deviation of offsets. However, the detected visible boxes with minor offsets still can be utilized to estimate the full body box precisely because of the robustness of FEN and EPM. Therefore, our method can work well although $\text{MR}^{-2}$ of visible box detection is higher than full box detection.

\begin{table}[ht]
	\centering
  \caption{Results of full-body box detection and visible box detection at different matching thresholds. The matching threshold means that at which level of IoU will a detection count as true positive. }
	\label{tbl:crowdhuman_iou}
	\footnotesize
	\begin{tabularx}{1.\linewidth}{X<{\centering}|X<{\centering}X<{\centering}|X<{\centering}X<{\centering}}
		\toprule
		& \multicolumn{2}{|c|}{Full box detection} & \multicolumn{2}{|c}{Visible box detection} \\
		\hline
		   IoU thres.& AP/\% & $\text{MR}^{-2}$/\%  &  AP/\% & $\text{MR}^{-2}$/\%  \\
		\hline
		0.5 & 85.18 & 49.18  &85.01 &53.75  \\
		\hline
		0.4  &88.68&44.23  &88.85&47.38\\
		\hline
		0.3  &90.76&41.89 &91.39&41.22  \\
		\hline
		0.2  &92.35&38.95  &93.20&35.08  \\
		\hline
		0.1 &93.82&35.85  &94.86&29.23  \\
		
		\bottomrule
	\end{tabularx}
\end{table}

Thanks to the aforementioned decomposition strategy, V2F-Net has the following advantages: (1) Unlike prior works tackle the above two tasks in one single network, each sub-network is responsible for its own task in V2F-Net. As a consequence, the learning of V2F-Net becomes simpler and can converge to better minimum. (2) The direct detection target is changed from full-body box to visible box, which can greatly reduce distraction of occluded region on features of pedestrian. (3) The intermediate product, visible box, can be utilized in NMS. By replacing \textit{Intersection over Union} (IoU) calculation on full-body boxes with visible boxes as~\cite{huang2020R2nms}, the dilemma for the single threshold of greedy-NMS can be eased a lot. 

In order to make the estimation of full-body box of pedestrian from visible box be more accurate, the features of visible region must contain enough information about human body parts, so that the network can know which direction to expand from the visible box and how much the offsets. Therefore, we propose a novel module that can perceive visibility of human body parts, called Embedding-based Part-aware Module (EPM). By adding a visibility loss for each divided part, the network is encouraged to extract features with essential part information. 


To summarize, our contributions are as follows:
\begin{itemize}
  \item We propose a simple yet effective pipeline to handle occlusion in pedestrian detection by explicit decomposition. It can be taken as a stronger baseline for occluded pedestrian detection.
 \item We propose a novel Embedding-based Part-aware Module (EPM) to further improve accuracy of full-body estimation. This module can be discarded during inference, thus will not bring extra computation cost.
  \item Our method improves the FPN baseline by 5.85\% AP on CrowdHuman and 2.24\% $\text{MR}^{-2}$ on CityPersons, achieving the state-of-the-art results on both the two challenging benchmarks. Besides, the consistent gains on both one-stage and two-stage detectors demonstrate the generalizability of our method.
\end{itemize}






\section{Related Work}
\paragraph{General Object Detection.}
CNN-based object detectors~\cite{ren2015faster, yolo, liu2016ssd, lin2017focal} have shown great superiority over methods using hand-crafted features~\cite{dollar2014fast, papageorgiou2000trainable}. The state-of-the-art detectors can be divided into two-stage methods and one-stage methods. Two-stage detectors~\cite{girshick2014rich, girshick2015fast, ren2015faster, cai2019cascadercnn} first generate a set of region proposals by methods like Selective Search~\cite{uijlings2013selective} and Region Proposal Network~\cite{ren2015faster}, then these proposals are fed into RCNN to do the final classification and localization. In contrast, one-stage detectors~\cite{yolov3, liu2016ssd, lin2017focal} directly predict objects based on dense sampling of possible locations, skipping the proposal stage. Generally speaking, two-stage detectors can achieve better accuracy while one-stage detectors have an advantage in computational efficiency.

\paragraph{Occluded Pedestrian Detection.}
Part-based approaches handle the occlusion by learning a series of part detectors at first, then fusing these detection results to generate final pedestrian boxes. Despite of effectiveness, it is time-consuming in inference phase. A few of previous works design novel loss functions without modifying the network architecture: Repulsion Loss~\cite{wang2018repulsion} takes consideration of the repulsion by other surrounding objects, in addition to the attraction by target; Aggregation Loss~\cite{zhang2018Occlusionaware} encourages proposals corresponding to the same pedestrian to be compact.

However, it is difficult for network to discriminate features between visible regions and occluded regions with annotated full-body box only. Therefore, some works try to leverage visible box to assist full-body pedestrian detection. Bi-box~\cite{Zhou_2018_bibox} is the first work to predict full-body box and visible box parallelly.~\cite{huang2020R2nms} improves it by proposal pairing.~\cite{zhang2018occludedattention} adds an extra occlusion pattern classification loss. These approaches enforce the detectors to focus on visible regions of pedestrians implicitly. Besides of them, there are also some works using an explicit way to achieve the same purpose. Feature re-weighting and re-scoring are two common strategies. Feature re-weighting methods re-weight features of pedestrian with information about visible parts:~\cite{zhang2018Occlusionaware} applies element-wise summation on RoI features of divided parts weighted by corresponding visibility scores;~\cite{pang2019mask} generates modulated features by a Mask-Guided Attention Branch, which takes RoI features and predicted spatial attention mask as input. In contrast, re-scoring approaches refine the pedestrian scores by scores of visible regions:~\cite{Zhou_2018_bibox} fuses scores of visible region and full body with softmax;~\cite{Noh_2018_CVPR} computes occlusion-aware detection score by applying a MLP layer to scores of parts.

Besides,~\cite{prnet} tackles occluded pedestrian detection as a progressive refinement process. It takes visible box as an auxiliary priori to initialize the full-body anchors, so as to build a fast one-stage detector. This is achieved through calibrating visible box anchors to a full-body template derived from occlusion statistics. Different from all these works, we propose to decompose occluded pedestrian detection into two more intuitive and simpler tasks. In the decomposed pipeline, the visible box can be fully taken advantage of to improve the detection performance.

\paragraph{NMS and its variants.}
NMS is adopted as a post-processing step of most object detectors to remove duplicate proposals belong to the same identity. In greedy-NMS, a proposal will be discarded if its IoU with more confident proposals is higher than the given IoU threshold. As we know, it will cause false suppression when using commonly used relatively low threshold in crowded scenes. Soft-NMS~\cite{softnms} improves it by leveraging a soft mechanism to decay the detection scores of pedestrian proposals, instead of eliminating them. However, only the locations and scores of full boxes are not enough to decide whether a proposal is redundant, thus can not adapt to complicate scenes well.

To handle this problem, some works propose to predict extra information by detection network in addition to locations and scores of object proposals, and utilize them in NMS. Adaptive-NMS~\cite{adaptiveNMS} outputs object density indicating the level of occluded occlusion. CaSe~\cite{xie2020count} predicts the number of pedestrians in the corresponding boxes and an embedding for discriminating different pedestrians in feature level. NOH-NMS~\cite{zhou2020NOH-NMS} introduces the nearby-objects distribution into the NMS pipeline. A more intuitive idea is using visible box to calculate IoU instead of full box. R$^2$NMS~\cite{huang2020R2nms} has shown its effectiveness on crowded pedestrian benchmarks. We also leverage this strategy to improve performance in our method.

\section{Motivation}
\begin{figure*}[!t]
\begin{center}
 \includegraphics[width=1.\linewidth]{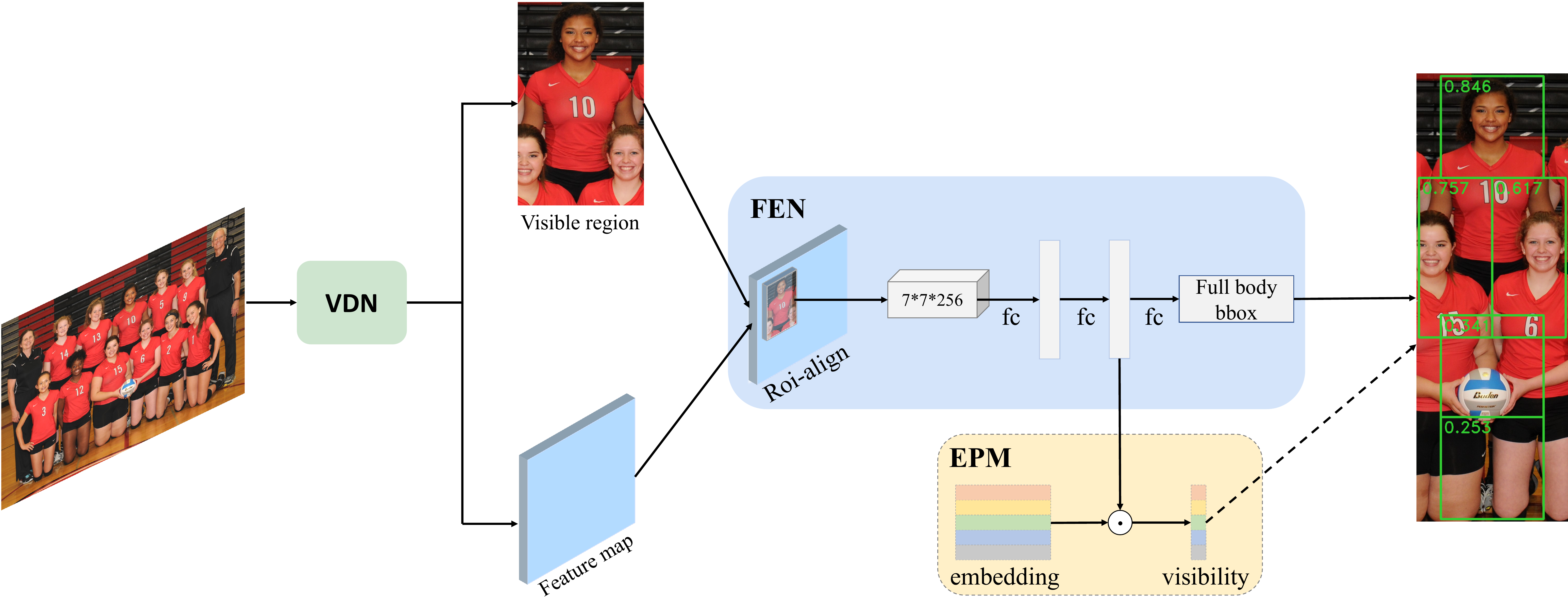}
\end{center}
   \caption{Framework of V2F-Net. The input image is first processed by Visible region Detection Network (VDN) to detect visible regions of all pedestrians. After NMS (only necessary during inference) these kept boxes are fed into Full body Estimation Network (FEN) to estimate the full body box for each pedestrian. During training, the visible boxes will also be passed to Embedding-based Part-aware Module (EPM) to predict visibility for each part of corresponding pedestrian. By supervising the part visibility, EPM works as an auxiliary module to make the estimation of full body be more accurate. $\bigodot$ is dot product operation. The dashed line and rectangle indicate they can be discarded during inference. In the output image, the green boxes and digits represent divided parts and  predicted scores by EPM respectively.}
\label{fig:method}
\end{figure*}

Given an image consists of occluded pedestrians like the input image in Fig.~\ref{fig:method}, it is hard to tell the precise full-body locations of them immediately even for human. Intuitively, the process of full-body pedestrian detection for human will go through two stages: in the first stage, we identify each pedestrian by its visible region; in the second stage, we estimate the full body box from the visible region. Human can ''see'' the invisible part because we have an empirical estimation about the structure and proportion of human body. However, most prior works fuse the detection task and estimation task into one single harder task, which would increase the difficulty of network learning.

A straightforward idea is to divide a hard problem and conquer each sub-problem separately. Such thought of decomposing has been proved to be effective in many computer vision tasks, e.g. pose regression~\cite{dollar2010cascaded_pose}, general object detection~\cite{cai2019cascadercnn} and face alignment~\cite{cao2014face,yan2013learn}. Inspired by these works, we propose to decompose occluded pedestrian detection into two sub-problems: visible region detection and full body estimation. The goal of our method is building a more intuitive and stronger pipeline to handle occlusion. In the decomposed pipeline, visible box can be fully taken advantage of to improve detection performance.

\section{Proposed Approach}

 In this section, we first clarify how V2F-Net decomposes occluded pedestrian detection into visible region detection and full body estimation. Then we introduce EPM to further improve accuracy of full-body box estimation from visible box. At last, we show how to train V2F-Net end-to-end.



\subsection{Pipeline of V2F-Net}
\label{sec:ppl}
Fig.~\ref{fig:method} presents the framework of our method. The V2F-Net consists of two sub-networks: Visible region Detection Network (VDN) and Full body Estimation Network (FEN), and an auxiliary module Embedding-based Part-aware Module (EPM). We will give a detailed introduction of them below. The pipeline of V2F-Net is straightforward during inference: an image is first processed by VDN to detect visible regions of all pedestrians. After NMS the kept visible boxes are fed into FEN to generate final full-body box for each pedestrian, no need for NMS anymore.

It is worth noting that we perform NMS only on the set of visible boxes for two reasons: (1) As suggested by~\cite{huang2020R2nms}, IoU between visible regions of two full-body boxes is a better indicator showing whether they belong to the same pedestrian, compared to using IoU between full-body boxes directly. Therefore, with an appropriate single IoU threshold, many duplicate pedestrian proposals can be removed. Meanwhile, both of boxes of two different pedestrians can be kept, even in crowded scenes. (2) After NMS on visible regions, only few of them can be passed to FEN, thus the FEN will not cost much time in inference phase. 

\paragraph{VDN.} The goal of VDN is to detect visible region of pedestrian. VDN can be implemented with minor modifications on the original detectors like Faster-RCNN~\cite{ren2015faster} and RetinaNet~\cite{lin2017focal}. The only thing we need to do is replacing the regression target from full box to visible box. Adjustment of anchors setting is optional, depending on whether anchor is necessary in chosen base detector. Many works utilize visible box to reduce interference of occluded regions on features of pedestrians by explicit attention~\cite{zhang2018occludedattention,pang2019mask} or implicit multi-task learning~\cite{Zhou_2018_bibox}. Compared to these methods, our solution is simpler and easier to implement.

The loss of VDN is the same as base detector, here we denote it by $\mathcal{L}_{VDN}$. With Faster-RCNN we have
\begin{equation}
\mathcal{L}_{VDN} = \mathcal{L}_{cls1}+\mathcal{L}_{reg1}+\mathcal{L}_{cls2}+\mathcal{L}_{reg2},
\label{equ:vdn_loss}
\end{equation}
    where $\mathcal{L}_{cls1}$, $\mathcal{L}_{reg1}$, $\mathcal{L}_{cls2}$ and $\mathcal{L}_{reg2}$ are classification loss and regression loss in RPN and RCNN, respectively.

\paragraph{FEN.} FEN aims to estimate the full body box from visible box that have already been detected by VDN. The architecture of FEN is almost the same as RCNN head in~\cite{ren2015faster}, except we replace RoI-Pooling with RoI-Align. Specifically, given visible box of a pedestrian \textit{v}, we use RoI-Align to extract corresponding features $\bm{F_v}$, then these features are fed into two consecutive FC layers with ReLU activation to transform $\bm{F_v}$ to more task-specific features $\bm{F_v^r}$, followed by another FC layer to predict full-body offsets from the input visible box.

To determine which full-body pedestrian box to predict from the input visible box \textit{v} during training, we use a \textit{vdt$\xrightarrow{}$vgt$\xrightarrow{}$fgt} label assignment strategy. To be formulated, let $\mathcal{G}={\{(v_i^*, f_i^*)|1 \leq i \leq N_g \}}$ be all ground truth pedestrians, the $i$-th pedestrian $\mathcal{G}_i$ is represented by its visible box $v_i^*$ and full box $f_i^*$ in a pair, $N_g$ is the total number of all these pedestrians. A visible box \textit{v} will be considered as negative if $\operatorname{IoU}(v, v_i^*)<0.5$ for all $i \in [1, N_g]$, otherwise it will be assigned to $\mathcal{G}_j$ as a positive sample, where $j=\operatorname*{argmax}_k\mathrm{IoU}(v, v_k^*), k\in{[1, N_g]}$. For these positive samples, we take the corresponding full box of assigned ground truth pedestrian as regression target. The loss of FEN $\mathcal{L}_{FEN}$ adopts similar formulation as regression loss of RCNN in Faster-RCNN, we refer to~\cite{ren2015faster} for more details.

\subsection{Embedding-based Part-aware Module}
The Embedding-based Part-aware Module is proposed to improve accuracy of full body estimation from visible region. At first, full body of pedestrian is divided into $n_p=5$ parts as~\cite{zhang2018Occlusionaware}, then we create a part embedding matrix $\bm{E}\in R^{n_p\times d_p}$, where $i$-th $d_p$-dimensional embedding $\bm{E_i}$ represents concept of $i$-th part $P_i$. The design of part embedding matrix is inspired by Word Embedding in NLP, where each word is represented by a vector to indicate its semantic information. Similarly, the part embedding matrix is shared across all data and trained with other parameters together. For discriminating
general concept of part and the part of a specific pedestrian, we use different notations, that is $P_i$ and $p_i$, to represent the $i$-th part.

Given the feature of detected visible region, we compute its response on each part to determine whether this part is visible in the given visible box. Specifically, we use inner product operation on $\bm{F_v^r}$ and $\bm{E}$, followed by sigmoid function to limit the response in range $(0, 1)$. Note that here we use the same feature $\bm{F_v^r}$ with FEN rather than transform $\bm{F_v}$ again, so that the gradients from the loss of EPM can back propagate through the FEN, guiding the regression of FEN to be more accurate. The above procedure can be formulated as:
\begin{equation}
r_i^p=\mathrm{sigmoid}(\bm{F_v^r}\cdot \bm{E_i}), i\in{[1, n_p]},
\label{equ:part_resp}
\end{equation}
where ``$\cdot$'' is inner product operation, and $r_i^p$ represents the response of feature $\bm{F_v^r}$ on part $p_i$. The higher the response $r_i^p$, the higher the visibility of part $p_i$. For each part, we use sigmoid cross-entropy loss to supervise the response to be close to corresponding ground truth label, enforcing the features of pedestrian to be aware of parts. The total loss of EPM is sum of losses for $n_p$ parts:
\begin{equation}
\mathcal{L}_{EPM} = \sum_{i=1}^{n_p}y_i^p \operatorname{log} r_i^p + (1-y_i^p) \operatorname{log} (1-r_i^p),
\label{equ:part_loss}
\end{equation}
    where $y_i^p$ represents visibility label for $i$-th part. $y_i^p$ is defined according to the IoA between input visible box \textit{v} and part $p_i$ of assigned pedestrian, we try different modes: hard label and soft label separately:
\begin{equation}
y_i^p =\left\{
             \begin{array}{lr}
             \mathbb{I}{(\operatorname{IoA}(v, p_i) \ge 0.5)}, & mode=hard \\
             \operatorname{IoA}(v, p_i), & mode=soft\\
             \end{array},
\right.
\label{equ: part_gt}
\end{equation}

\begin{equation}
    \operatorname{IoA}(a, b) = \frac{\operatorname{area}(a \cap b)}{\operatorname{area}(a)},
\label{equ: ioa}
\end{equation}
where $\mathbb{I}{(.)}$ is an indicator function. We experimentally find the hard way performs slightly better than the soft , so we use hard label in this paper unless otherwise specified. Note that EPM only works in training phase, so no extra computation cost is brought by this module during inference.

\subsection{Training}
Although we decompose occluded pedestrian detection into two successive sub-problems, V2F-Net still can be trained end-to-end. We omit the training of VDN as there is no difference between base detector and VDN.

We start with the visible boxes detected by VDN $V$. In case of too few positive samples for FEN and EPM, we skip NMS on these visible boxes during training and enrich the training set with ground truth visible boxes. We denote the augmented set of visible boxes as $\mathcal{V}=V \cup \mathcal{G}^v$, where $\mathcal{G}^v=\{v_i^*|i\in[1, N_g]\}$ represents visible boxes of all ground truth pedestrians. After label assignment as Sec.~\ref{sec:ppl}, we randomly sample $\operatorname{max}(1000, |\mathcal{V}|)$ visible boxes from $\mathcal{V}$ by ratio positive:negative=9:1. All these samples are fed into EPM, while only positive samples are passed to FEN. The total loss of V2F-Net $\mathcal{L}$ is weighted sum of above three losses:

\begin{equation}
\mathcal{L} = \mathcal{L}_{VDN}+\alpha\mathcal{L}_{FEN}+\beta\mathcal{L}_{EPM},
\label{equ: total_loss}
\end{equation}
where $\alpha$, $\beta$ are balanced factors of $\mathcal{L}_{FEN}$ and $\mathcal{L}_{EPM}$, respectively.




\section{Experiments}
To verify the effectiveness of our method, we conduct experiments on two standard crowded datasets: CrowdHuman~\cite{shao2018crowdhuman} and CityPersons~\cite{zhang2017citypersons}. Besides comparison of quantified results, we also visualize the learned part visibility to prove that EPM works as we expected. At last, we give some analysis and discussions about how to further improve V2F-Net.

\subsection{Implementation Details}
We use Faster-RCNN~\cite{ren2015faster} with FPN~\cite{lin2017feature} as our baseline. RoI-Pooling is replaced with RoI-Align~\cite{he2017mask}, the backbone is ResNet-50 pre-trained on ImageNet~\cite{russakovsky2015imagenet}. The anchor setting is the same as~\cite{wang2018repulsion,chu2020crowddet}, which uses scale \{1\} and ratio $H/W=\{1, 2, 3\}$ for both CrowdHuman and CityPersons. As for our method, we use the same anchor scale with baseline, but change ratio to \{0.5, 1, 1.5\} for visible box detection. All the models are trained with batch of 16 images on 8 GPUs and  optimized by Stochastic Gradient Descent (SGD) with 0.9 momentum, the weight decay is set to 0.0001. IoU threshold in NMS during inference is set to 0.5, regardless of whether NMS is performed on visible boxes or full boxes. All added FC layers are initialized following~\cite{lin2017feature}. Each element in Part Embedding Matrix $E$ is randomly sampled from uniform distribution with range [-0.0005, 0.0005]. For all experiments in this paper, we use AP and $\text{MR}^{-2}$ as evaluation metrics. Higher AP indicate better performance, while $\text{MR}^{-2}$ is the opposite.

For CrowdHuman, we resize image to make sure the shorter edge equals to 800 pixels, while the longer edge is smaller than 1400 pixels during both training and inference. We train 30 epoches with initial learning rate 0.00125. The learning rate will be decayed by factor of 0.1 at epoch 24 and 27. By default we use $\alpha=0.3$, $\beta=1$.

For CityPersons, the image size is upsampled by 1.3x in both training and test following~\cite{zhang2017citypersons}. We train our models on all training set for 35 epoches. We set the initial learning rate to 0.002, then decay it to 0.0002 at 20th epoch and 0.00002 at 30th epoch. The balanced factors of losses are set to $\alpha=0.5$, $\beta=1$.

\subsection{Experiments on CrowdHuman}
\label{sec:exp_crowdhuman}
CrowdHuman~\cite{shao2018crowdhuman} is a recently released dataset for better evaluating various pedestrian detectors in crowded scenes. On the average, there are 22.6 pedestrians in an image and 2.4 pedestrians have IoU$>$0.5 with other pedestrians. This dataset is split into training set, validation set and test set, each of them contains 15000, 4370, 5000 images respectively. All the models in this paper are trained on the training set and evaluated on validation set.

\paragraph{Ablation study.} Table~\ref{tbl:crowd_ablation} shows the ablation results on CrowdHuman validation set. Our re-implemented baseline is better than results in original paper~\cite{shao2018crowdhuman}. Obviously, both the decomposed pipeline V2F and the auxiliary module EPM can improve performance of full-body pedestrian detection. By simply modifying the pipeline from detecting full body of pedestrian directly, to do visible region detection and full body estimation sequentially, we achieve 5.63\% AP and 3.89\% $\text{MR}^{-2}$ gains. Based on the decomposed V2F pipeline, the cost-free EPM can further improve 0.22\% AP and 0.78\% $\text{MR}^{-2}$. The results validate the effectiveness of our proposed method. Not only the recall can be improved by a large margin, but also it will not bring more false positives.

\begin{table}[ht]
	\centering
  \caption{ Ablation experiments conducted on CrowdHuman validation set. \emph{V2F} indicates our proposed pipeline which do visible region detection and full body estimation sequentially. \emph{EPM} is the Embedding-based Part-aware Module. Best results are boldfaced. 
	}
	\footnotesize
	\label{tbl:crowd_ablation}
	\begin{tabularx}{1.\linewidth}{c|cc|X<{\centering}X<{\centering}X<{\centering}}
		\toprule
		  Method & V2F & EPM & AP/\% & $\text{MR}^{-2}$/\% & Recall/\% \\
		\hline
		baseline in~\cite{shao2018crowdhuman} & & & 84.95 & 50.42  & ---\\
		our baseline & & & 85.18 & 46.95 & 76.90 \\
		\hline
		\multirow{2}*{V2F-Net} & \checkmark & & 90.81 & 43.06   & 83.92\\
		& \checkmark & \checkmark & \textbf{91.03} & \textbf{42.28}  & \textbf{84.20}\\
		\bottomrule
	\end{tabularx}

\end{table}

\paragraph{Different pipelines for full-body pedestrian detection.} In addition to our proposed pipeline \emph{V2F}, there are two strategies for generating full-body box of a pedestrian: detect full body pedestrian only, and predict full body box and visible box parallelly like~\cite{Zhou_2018_bibox}. We denote them by \emph{F} and \emph{V\&F}, respectively. One may doubt that the improvement of \emph{V2F} is because of extra computation cost brought by the FEN. Therefore, we add an extra RCNN head with the same architecture as FEN to our baseline following \textit{iterative bounding box regression} as~\cite{gidaris2015object,Gidaris2016Attend}. Here we denote this strategy as \emph{$F^2$}. For fair comparison, we do not add classification branch for re-scoring in \emph{F$^2$}, which means the second RCNN is only used to refine locations of full-body boxes detected by the first one.

Table~\ref{tbl:crowd_ablation_ppl} presents the results of aforementioned pipelines on CrowdHuman validation set. We can draw the following conclusions from it: (1) By just adding an extra task of visible region detection, the performance of full-body pedestrian detection can be improved by 1.78\% $\text{MR}^{-2}$ , which is consistent with the conclusion in~\cite{Zhou_2018_bibox}. (2) The mode of iterative regression has a positive effect indeed, but still 3.3\% AP and 0.83\% $\text{MR}^{-2}$ worse than our decomposed pipeline when using visible boxes for NMS. (3) Replacing calculation of IoU on full-body boxes with visible boxes can not guarantee to bring improvements. It works only when the visible boxes are precise enough (see Table~\ref{tbl:crowd_ablation_ppl_v} for quantitative results). (4) Our decomposed pipeline beats all others by 3.3\%$ \sim $5.63\% AP and 0.83\%$ \sim $3.89\% $\text{MR}^{-2}$. In conclusion, our proposed pipeline can be taken as a stronger baseline for occluded pedestrian detection.

\begin{table}[ht]
	\centering
  \caption{Results of different pipelines for full-body pedestrian detection on CrowdHuman validation set. In the first column, \emph{F} means detecting full-body pedestrian only, \emph{V\&F} means predicting visible box and full box parallelly, \emph{F$^2$} means iterative full body regression by 2 steps, \emph{V2F} is our proposed method without EPM. The second column indicates the inputs of NMS are visible boxes or full body boxes.}
	\label{tbl:crowd_ablation_ppl}
	\footnotesize
	\begin{tabularx}{1.\linewidth}{X<{\centering}|X<{\centering}|X<{\centering}X<{\centering}X<{\centering}}
		\toprule
		  Pipeline & NMS & AP/\% & $\text{MR}^{-2}$/\%  & Recall/\% \\
		\hline
		F & full & 85.18 & 46.95  & 76.90 \\
		\hline
		\multirow{2}*{V\&F} & full & 85.19 & 45.17  & 77.71  \\
		& visible & 86.70 & 51.94  & 79.61  \\
		\hline
		F$^2$ & full & 87.51 & 43.89  & 79.96 \\
		\hline
		\multirow{2}*{V2F} & full & 86.04 & 43.84  &80.20\\
		& visible & \textbf{90.81} & \textbf{43.06}   & \textbf{83.92} \\
		\bottomrule
	\end{tabularx}
\end{table}

\begin{table}[ht]
  \caption{Evaluation of visible region detection for different pipelines. Ours is much better than result of \emph{V\&F}.}
	\label{tbl:crowd_ablation_ppl_v}
	\centering
	\footnotesize
	\begin{tabular}{c|ccc}
		\toprule
		  Pipeline & AP/\% & $\text{MR}^{-2}$/\% & Recall/\%\\
		\hline
		V\&F & 78.92 & 65.78 & 74.78  \\
		\hline
		V2F & \textbf{84.90} & \textbf{51.93} & \textbf{78.54}\\
		\bottomrule
	\end{tabular}
\end{table}

\paragraph{Generalizability of V2F-Net.} In V2F-Net, the VDN can be implemented with minor modifications based on chosen detector. The FEN and EPM only take the multi-scale feature maps and detected visible boxes from VDN as inputs. Therefore, theoretically both one-stage and two-stage detectors can be incorporated into V2F-Net. To demonstrate this, we take Faster-RCNN~\cite{ren2015faster} and RetinaNet~\cite{lin2017focal} as the representative of the two types of detectors respectively, and implement our method based on them. Both the two methods utilize FPN~\cite{lin2017feature}. The implementation details of RetinaNet are almost the same as original paper, except we use anchor ratios \{1, 2, 3\} for better performance. From Table~\ref{tbl:crowd_ablation_detector} we can see that V2F-Net can achieve consistent gains when taking Faster-RCNN or RetinaNet as base detector. The results validate the generalizability of our method. 

\begin{table}[ht]
	\centering
    \caption{Comparison of results when using different detectors with/without V2F-Net. Both the Faster-RCNN and RetinaNet utilize FPN~\cite{lin2017feature} for better performance.}
	\label{tbl:crowd_ablation_detector}
	\begin{tabularx}{1.\linewidth}{c|X<{\centering}|X<{\centering}X<{\centering}X<{\centering}}
		\toprule
		  Detector & Method & AP & $\text{MR}^{-2}$ & Recall \\
		\hline
		\multirow{2}*{FRCNN~\cite{ren2015faster}} & baseline & 85.18 & 46.95 & 76.90 \\
		& ours &  \textbf{91.03}  &  \textbf{42.28} &  \textbf{84.20} \\
		\hline
		\multirow{2}*{RetinaNet~\cite{lin2017focal}} & baseline & 81.81 & 56.64 & 74.58 \\
		& ours & \textbf{84.92} &  \textbf{53.99} &  \textbf{76.75}\\
		\bottomrule
	\end{tabularx}
\end{table}

\paragraph{Hard label \vs~Soft label in EPM.} We compare the results between hard label and soft label for each divided part in Table~\ref{tbl:crowd_label_EPM}. The hard way performs slightly better than the soft. We suspect that is because the adopted strategy of dividing part does not consider the human pose, causing the soft label increases the ambiguity of part visibility instead.

\begin{table}[ht]
	\centering
  \caption{Comparison of different modes for part label in EPM. All the other settings are the same.
	}
	\label{tbl:crowd_label_EPM}
	\begin{tabular}{c|cc}
		\toprule
		 Mode & AP/\% & $\text{MR}^{-2}$/\% \\
		\hline
		hard & \textbf{91.03} & \textbf{42.28} \\
		soft & 90.35 & 42.67 \\
		\bottomrule
	\end{tabular}
\end{table}

\begin{figure}[!t]
\begin{center}
 \includegraphics[width=1.\linewidth]{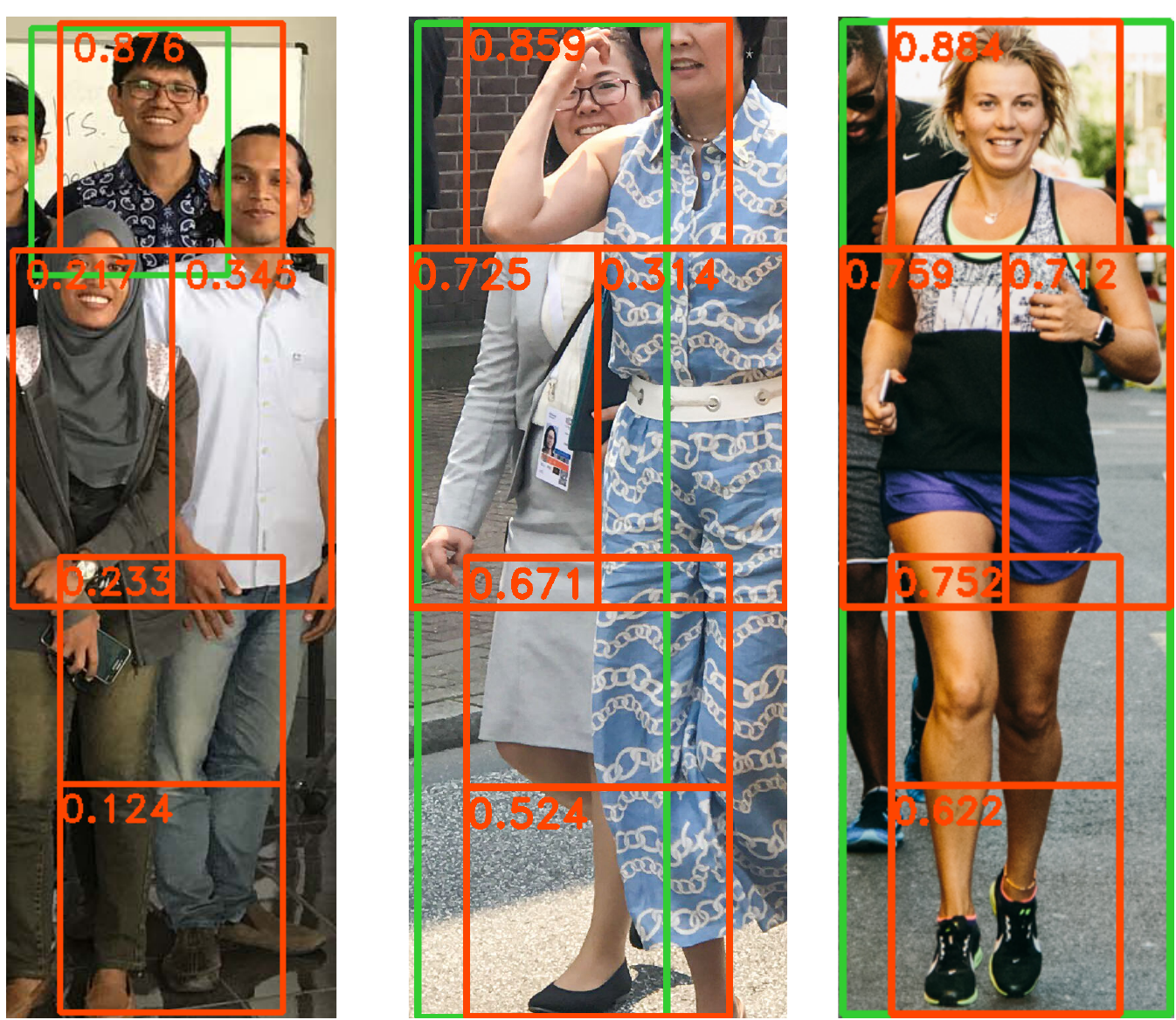}
\end{center}
   \caption{Visualization of predicted scores for each part by EPM. The example images are cropped from original images according to detected full-body box of pedestrian. The green and red rectangles represent detected visible boxes and five parts divided as~\cite{zhang2018Occlusionaware}. Each predicted score for a part indicates its visibility. Best viewed in color.}
\label{fig:vis_part}
\end{figure}

\paragraph{Visualization of EPM.} To demonstrate the EPM works as expected, we visualize the scores predicted by EPM for each divided part. Fig.~\ref{fig:vis_part} presents some examples. As we can see, there is a roughly positive correlation between the score and visibility for each part.

\paragraph{Comparison with the state-of-the-art methods.} We show the comparison between our method and the state-of-the-art methods on CrowdHuman validation set in Table~\ref{tbl:crowdhuman_eval}. All methods take the same backbone (Res-50 + FPN), and are evaluated using the same image size. Our method outperforms most of the state-of-the-art pedestrian detectors, but slightly worse than CrowdDet~\cite{chu2020crowddet} due to our relatively weaker baseline result. In Sec.~\ref{sec:discussion} we will discuss about the combination of these SOTA methods and V2F-Net.
\begin{table}[ht]
   \centering
   \footnotesize
   \caption{Comparison of various crowded detection methods on CrowdHuman validation set. All methods employee the FPN with Res-50 backbone as baseline, and are evaluated using the same image size.}
   \label{tbl:crowdhuman_eval}
   \begin{tabularx}{1.\linewidth}{p{40mm}<{\centering}|X<{\centering}X<{\centering}X<{\centering}}
   \toprule
       Method & AP/\% & $\text{MR}^{-2}$/\% & Recall/\%\\
       \hline
        Baseline & 85.18 & 46.95 & 76.90 \\
       \hline
       Adaptive-NMS~\cite{adaptiveNMS} & 84.71 & 49.73 & --- \\
       R$^2$NMS~\cite{huang2020R2nms} & 89.29 & 43.35 & --- \\
       CaSe~\cite{xie2020count} & --- & 47.9 & ---\\
       NOH-NMS~\cite{zhou2020NOH-NMS} & 89.0 & 43.9 & ---\\
       CrowdDet~\cite{chu2020crowddet} & 90.7 & 41.4 & 83.68 \\
       \hline
       Ours & \textbf{91.03} & 42.28  & \textbf{84.20}\\
       \bottomrule
   \end{tabularx}
\end{table}

\begin{figure*}[!t]
\begin{center}
 \includegraphics[width=1.\linewidth]{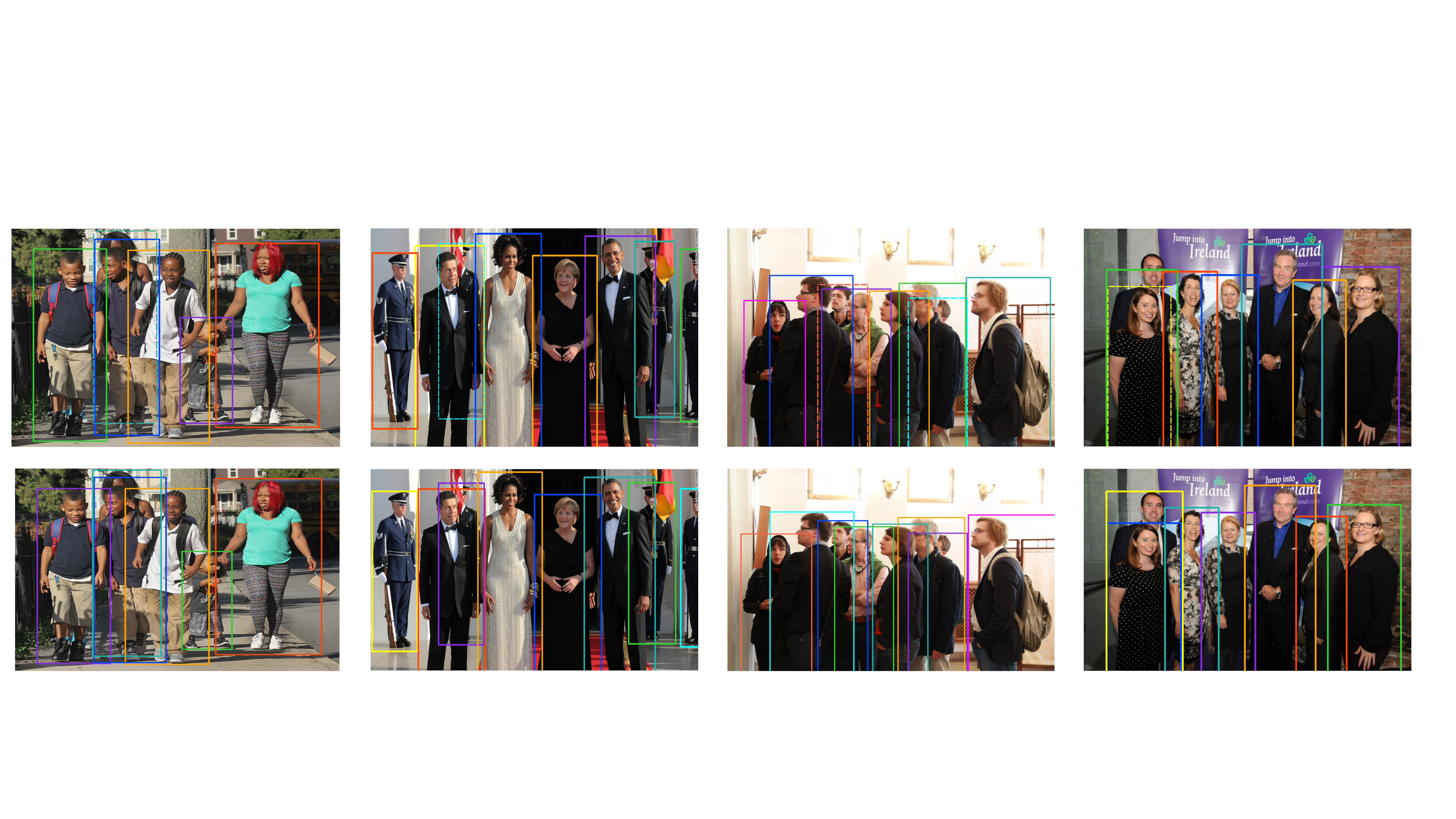}
\end{center}
   \caption{Visualization of detection results. The first row come from FPN baseline and the second are our results. All detected boxes are filtered by score$\ge$0.3. The boxes in solid and dashed line represent kept boxes and false suppression by NMS respectively.}
\label{fig:example_cropped}
\end{figure*}

\subsection{Experiments on CityPersons}
CityPersons~\cite{zhang2017citypersons} is a subset of CityScapes~\cite{cordts2016cityscapes} which focus on pedestrian detection. There are 2975, 500, 1575 images for training, validation and testing, respectively. Compared to CrowdHuman, CityPersons is less crowded, but still contains lots of occluded cases. We train our models on all training set, and report results on reasonable validation subset.

\paragraph{Comparison with the state-of-the-art methods.} Table~\ref{tbl:citypersons_eval} lists results of the state-of-the-art methods and V2F-Net on CityPersons. Similar to the results on CrowdHuman, both the decomposed strategy and EPM can consistently improve the detection performance. Our method achieves 2.24\% $\text{MR}^{-2}$ and 0.94\% AP gains compared to our FPN baseline. The improved result is comparable or even better than the state-of-the-art methods.

\begin{table}[ht]
   \centering
   \caption{Comparison between the state-of-the-art methods and ours on CityPersons validation set. The third column indicates the enlarge number of original image in both training and testing. \emph{Ours-V2F} is the simplified version of our V2F-Net without EPM.}
   \label{tbl:citypersons_eval}
   \begin{tabular}{c|c|c|cc}
   \toprule
   Method & Backbone & Scale & $\text{MR}^{-2}$ & AP \\
   \hline
   Baseline & Res-50 & $\times$1.3 & 12.32 & 95.25  \\
   \hline
   AF-RCNN~\cite{zhang2017citypersons} & \multirow{9}*{VGG-16} & $\times$1.3 & 12.81 & ---\\
   OR-CNN~\cite{zhang2018Occlusionaware} &  & $\times$1.3 & 11.0 & ---\\
   FRCN~\cite{zhou2019discriminative} &  & $\times$1.3 & 11.1 & --- \\
   Adaptive~\cite{adaptiveNMS}  & & $\times$1.3 & 10.8 & ---  \\
   MGAN~\cite{pang2019mask} & & $\times$1.3 & 10.5 & --- \\
   GA~\cite{zhang2018occludedattention} & & $\times$1 & 15.96 & --- \\
   Bi-box~\cite{Zhou_2018_bibox} & & $\times$1.3 & 11.24 & --- \\
   R$^2$NMS~\cite{huang2020R2nms} & & $\times$1 & 11.1 & --- \\
   \hline
   Repulsion~\cite{wang2018repulsion}& \multirow{4}*{Res-50} & $\times$1.3 & 11.6 & --- \\
   ALFNet~\cite{liu2018ALFNet} & & $\times$1 & 12.0 & --- \\
   CrowdDet~\cite{chu2020crowddet} & & $\times$1.3 & 10.7 & 96.1  \\
   PRNet~\cite{prnet} & & $\times$1 & 10.8 & ---  \\
   \hline
   Ours-V2F & \multirow{2}*{Res-50} & $\times$1.3 & 11.33 & 95.89 \\
   Ours & & $\times$1.3 & \textbf{10.08} & \textbf{96.19} \\
   \bottomrule
   \end{tabular}
\end{table}

\subsection{Discussion}
\label{sec:discussion}
Although our method has achieved an inspiring improvement, we believe the V2F-Net is just a new baseline based on the effective decomposed pipeline. Here we will make a discussion about our method, aiming to give some inspiration to the following works.

\paragraph{Performance Analysis.} Thanks to the decomposed solution proposed in this paper, we can do more detailed analysis to guide how to optimize it further. To explore the overall performance limited by each component in V2F-Net, we conduct three experiments, assuming the VDN, VDN+NMS or FEN to be perfect respectively: (1) \emph{P-VDN}: replace the boxes detected by VDN with the ground truth visible boxes $\mathcal{G}^v$. (2) \emph{P-VDN+NMS}: take $\mathcal{G}^v$ as inputs of FEN. (3) \emph{P-FEN}: the full box of each detected visible boxes are obtained by label assignment as training, instead of estimated by FEN. The difference between the first two experiments is whether to do NMS on $\mathcal{G}^v$. Setting the scores of these ground truth visible boxes equally will confuse the NMS and evaluation metrics like AP and $\text{MR}^{-2}$, as they require the input boxes are sorted by confidence. Actually we tried this strategy and found it will cause $\text{MR}^{-2}$ increasing rapidly. Therefore, we predict scores for ground truth boxes by feeding them into VDN.

From Table~\ref{tbl:upper-analysis} we find the performance of above experiments showing the same trend on both CrowdHuman and CityPersons. Both \emph{P-VDN} and \emph{P-FEN} improve AP slightly but reduce $\text{MR}^{-2}$ by a large margin. This indicates VDN produces many false positives with high score, and FEN should pay more attention to these highly confident false positives visible boxes. Compared to \emph{P-VDN}, \emph{P-VDN+NMS} improves AP and $\text{MR}^{-2}$ to some extent, showing that there is still false suppression even using visible boxes in NMS.

\begin{table}[ht]
	\centering
  \caption{Qualitative analysis about the detection performance limited by each component in V2F-Net. \emph{P-VDN}, \emph{P-VDN+NMS} and \emph{P-FEN} indicate upgraded V2F-Net using perfect VDN, VDN+NMS and FEN respectively, which are obtained by ``cheating'' with ground truth boxes.
	}
	\label{tbl:upper-analysis}
	\begin{tabularx}{1.\linewidth}{c|c|X<{\centering}X<{\centering}}
		\toprule
		Dataset & Method & AP/\% & $\text{MR}^{-2}$/\% \\
		\hline
		\multirow{4}*{CrowdHuman} & V2F-Net & 91.03 & 42.31 \\
    & P-VDN & 91.99 & 20.52 \\
		& P-VDN+NMS & 95.14 & 17.67 \\
		& P-FEN & 92.25 & 35.56 \\
		\hline
		\multirow{4}*{CityPersons} & V2F-Net & 96.19 & 10.08 \\
		& P-VDN & 96.83 & 0.0310 \\
		& P-VDN+NMS & 99.68 & 0.0025 \\
		& P-FEN & 97.37 & 0.0929 \\
		\bottomrule
	\end{tabularx}
\end{table}


\paragraph{Beyond V2F-Net.} Despite the effectiveness, V2F-Net still suffer from some common failure cases as other pipelines: crowd errors, false suppression by NMS, etc. Fortunately, many excellent works have proposed effective approaches to solve these problems, e.g.~\cite{chu2020crowddet} for label assignment ambiguity,~\cite{wang2018repulsion,zhang2018Occlusionaware} for crowd error,~\cite{adaptiveNMS, Noh_2018_CVPR, xie2020count} for NMS. We believe the performance of V2F-Net can be much better when combined with these brilliant ideas.

\section{Conclusion}
We propose a simple yet effective method to handle occlusion in pedestrian detection: V2F-Net. By decomposing occluded pedestrian detection into visible region detection and full body estimation, the learning of network becomes easier and can converge to better minimum. To further improve the accuracy of full body estimation, we propose a novel module called EPM, which is cost-free during inference. We experimentally show the effectiveness of the decomposed pipeline and EPM, and validate the generalizability of our method on both one-stage and two-stage detectors. We consider V2F-Net as a new baseline for occluded pedestrian detection. And we believe when combined with other brilliant ideas, the evolution of this pipeline will lead to much better performance beyond ours.

{\small
\bibliographystyle{ieee_fullname}
\bibliography{v2fnet}

\begin{thebibliography}{10}\itemsep=-1pt

\bibitem{softnms}
Navaneeth Bodla, Bharat Singh, Rama Chellappa, and Larry~S. Davis.
\newblock Soft-nms -- improving object detection with one line of code.
\newblock In {\em Proceedings of the IEEE International Conference on Computer
  Vision (ICCV)}, pages 5561--5569, Oct 2017.

\bibitem{cai2019cascadercnn}
Zhaowei Cai and Nuno Vasconcelos.
\newblock Cascade r-cnn: High quality object detection and instance
  segmentation.
\newblock {\em IEEE Transactions on Pattern Analysis and Machine Intelligence},
  2019.

\bibitem{cao2014face}
Xudong Cao, Yichen Wei, Fang Wen, and Jian Sun.
\newblock Face alignment by explicit shape regression.
\newblock {\em International Journal of Computer Vision}, 107(2):177--190,
  2014.

\bibitem{chi2019pedhunter}
Cheng Chi, Shifeng Zhang, Junliang Xing, Zhen Lei, Stan~Z Li, and Xudong Zou.
\newblock Pedhunter: Occlusion robust pedestrian detector in crowded scenes.
\newblock In {\em Proceedings of the AAAI Conference on Artificial
  Intelligence}, pages 10639--10646, 2020.

\bibitem{chu2020crowddet}
Xuangeng Chu, Anlin Zheng, Xiangyu Zhang, and Jian Sun.
\newblock Detection in crowded scenes: One proposal, multiple predictions.
\newblock In {\em Proceedings of the IEEE Conference on Computer Vision and
  Pattern Recognition (CVPR)}, pages 12214--12223, 2020.

\bibitem{cordts2016cityscapes}
Marius Cordts, Mohamed Omran, Sebastian Ramos, Timo Rehfeld, Markus Enzweiler,
  Rodrigo Benenson, Uwe Franke, Stefan Roth, and Bernt Schiele.
\newblock The cityscapes dataset for semantic urban scene understanding.
\newblock In {\em Proceedings of the IEEE conference on Computer Vision and
  Pattern Recognition (CVPR)}, pages 3213--3223, 2016.

\bibitem{dai2016rfcn}
Jifeng Dai, Yi Li, Kaiming He, and Jian Sun.
\newblock R-fcn: Object detection via region-based fully convolutional
  networks.
\newblock In {\em Advances in Neural Information Processing Systems (NeurIPS)},
  pages 379--387, 2016.

\bibitem{dollar2014fast}
Piotr Doll{\'a}r, Ron Appel, Serge Belongie, and Pietro Perona.
\newblock Fast feature pyramids for object detection.
\newblock {\em IEEE Transactions on Pattern Analysis and Machine Intelligence},
  36(8):1532--1545, 2014.

\bibitem{dollar2010cascaded_pose}
Piotr Doll{\'a}r, Peter Welinder, and Pietro Perona.
\newblock Cascaded pose regression.
\newblock In {\em Proceedings of the IEEE Conference on Computer Vision and
  Pattern Recognition (CVPR)}, pages 1078--1085, 2010.

\bibitem{gidaris2015object}
Spyros Gidaris and Nikos Komodakis.
\newblock Object detection via a multi-region and semantic segmentation-aware
  cnn model.
\newblock In {\em Proceedings of the IEEE International Conference on Computer
  Vision (ICCV)}, pages 1134--1142, 2015.

\bibitem{Gidaris2016Attend}
Spyros Gidaris and Nikos Komodakis.
\newblock Attend refine repeat: Active box proposal generation via in-out
  localization.
\newblock In {\em Proceedings of the British Machine Vision Conference (BMVC)},
  pages 90.1--90.13, 2016.

\bibitem{girshick2015fast}
Ross Girshick.
\newblock Fast r-cnn.
\newblock In {\em Proceedings of the IEEE International Conference on Computer
  Vision (ICCV)}, pages 1440--1448, 2015.

\bibitem{girshick2014rich}
Ross Girshick, Jeff Donahue, Trevor Darrell, and Jitendra Malik.
\newblock Rich feature hierarchies for accurate object detection and semantic
  segmentation.
\newblock In {\em Proceedings of the IEEE Conference on Computer Vision and
  Pattern Recognition (CVPR)}, pages 580--587, 2014.

\bibitem{he2017mask}
Kaiming He, Georgia Gkioxari, Piotr Dollár, and Ross Girshick.
\newblock Mask {R-CNN}.
\newblock In {\em Proceedings of the IEEE International Conference on Computer
  Vision (ICCV)}, pages 2980--2988, 2017.

\bibitem{huang2020R2nms}
Xin Huang, Zheng Ge, Zequn Jie, and Osamu Yoshie.
\newblock Nms by representative region: Towards crowded pedestrian detection by
  proposal pairing.
\newblock In {\em Proceedings of the IEEE Conference on Computer Vision and
  Pattern Recognition (CVPR)}, pages 10750--10759, 2020.

\bibitem{lin2017feature}
Tsung-Yi Lin, Piotr Doll{\'a}r, Ross Girshick, Kaiming He, Bharath Hariharan,
  and Serge Belongie.
\newblock Feature pyramid networks for object detection.
\newblock In {\em Proceedings of the IEEE Conference on Computer Vision and
  Pattern Recognition (CVPR)}, page~4, 2017.

\bibitem{lin2017focal}
Tsung-Yi Lin, Priya Goyal, Ross Girshick, Kaiming He, and Piotr Doll{\'a}r.
\newblock Focal loss for dense object detection.
\newblock In {\em Proceedings of the IEEE International Conference on Computer
  Vision (CVPR)}, pages 2980--2988, 2017.

\bibitem{adaptiveNMS}
Songtao Liu, Di Huang, and Yunhong Wang.
\newblock Adaptive nms: Refining pedestrian detection in a crowd.
\newblock In {\em Proceedings of the IEEE Conference on Computer Vision and
  Pattern Recognition (CVPR)}, pages 6459--6468, 2019.

\bibitem{liu2016ssd}
Wei Liu, Dragomir Anguelov, Dumitru Erhan, Christian Szegedy, Scott Reed,
  Cheng-Yang Fu, and Alexander~C. Berg.
\newblock {SSD}: Single shot multibox detector.
\newblock In {\em Proceedings of the European Conference on Computer Vision
  (ECCV)}, pages 21--37, 2016.

\bibitem{liu2018ALFNet}
Wei Liu, Shengcai Liao, Weidong Hu, Xuezhi Liang, and Xiao Chen.
\newblock Learning efficient single-stage pedestrian detectors by asymptotic
  localization fitting.
\newblock In {\em Proceedings of the European Conference on Computer Vision
  (ECCV)}, pages 618--634, 2018.

\bibitem{Noh_2018_CVPR}
Junhyug Noh, Soochan Lee, Beomsu Kim, and Gunhee Kim.
\newblock Improving occlusion and hard negative handling for single-stage
  pedestrian detectors.
\newblock In {\em Proceedings of the IEEE Conference on Computer Vision and
  Pattern Recognition (CVPR)}, pages 966--974, 2018.

\bibitem{discriminativeoccluding}
Wanli Ouyang and Xiaogang Wang.
\newblock A discriminative deep model for pedestrian detection with occlusion
  handling.
\newblock In {\em Proceedings of the IEEE Conference on Computer Vision and
  Pattern Recognition (CVPR)}, pages 3258--3265, June 2012.

\bibitem{pang2019mask}
Yanwei Pang, Jin Xie, Muhammad~Haris Khan, Rao~Muhammad Anwer, Fahad~Shahbaz
  Khan, and Ling Shao.
\newblock Mask-guided attention network for occluded pedestrian detection.
\newblock In {\em Proceedings of the IEEE International Conference on Computer
  Vision (ICCV)}, pages 4967--4975, 2019.

\bibitem{papageorgiou2000trainable}
Constantine Papageorgiou and Tomaso Poggio.
\newblock A trainable system for object detection.
\newblock {\em International journal of computer vision}, 38(1):15--33, 2000.

\bibitem{yolo}
Joseph Redmon, Santosh Divvala, Ross Girshick, and Ali Farhadi.
\newblock You only look once: Unified, real-time object detection.
\newblock In {\em Proceedings of the IEEE Conference on Computer Vision and
  Pattern Recognition (CVPR)}, pages 779--788, June 2016.

\bibitem{yolov3}
Joseph Redmon and Ali Farhadi.
\newblock Yolov3: An incremental improvement.
\newblock Technical report, 2018.

\bibitem{ren2015faster}
Shaoqing Ren, Kaiming He, Ross Girshick, and Jian Sun.
\newblock Faster {R-CNN}: Towards real-time object detection with region
  proposal networks.
\newblock In {\em Advances in Neural Information Processing Systems
  ({NeurIPS})}, pages 91--99, 2015.

\bibitem{optimizedpedestrian}
Sitapa Rujikietgumjorn and Robert~T. Collins.
\newblock Optimized pedestrian detection for multiple and occluded people.
\newblock In {\em Proceedings of the IEEE Conference on Computer Vision and
  Pattern Recognition (CVPR)}, pages 3690--3697, June 2013.

\bibitem{russakovsky2015imagenet}
Olga Russakovsky, Jia Deng, Hao Su, Jonathan Krause, Sanjeev Satheesh, Sean Ma,
  Zhiheng Huang, Andrej Karpathy, Aditya Khosla, Michael Bernstein, et~al.
\newblock Imagenet large scale visual recognition challenge.
\newblock {\em International Journal of Computer Vision}, 115(3):211--252,
  2015.

\bibitem{shao2018crowdhuman}
Shuai Shao, Zijian Zhao, Boxun Li, Tete Xiao, Gang Yu, Xiangyu Zhang, and Jian
  Sun.
\newblock Crowdhuman: A benchmark for detecting human in a crowd.
\newblock {\em arXiv preprint arXiv:1805.00123}, 2018.

\bibitem{prnet}
Xiaolin Song, Kaili Zhao, Wen-Sheng Chu, Honggang Zhang, and Jun Guo.
\newblock Progressive refinement network for occluded pedestrian detection.
\newblock In {\em Proceedings of the European Conference on Computer Vision
  (ECCV)}, page~9, September 2020.

\bibitem{stewart2016end}
Russell Stewart, Mykhaylo Andriluka, and Andrew~Y Ng.
\newblock End-to-end people detection in crowded scenes.
\newblock In {\em Proceedings of the IEEE Conference on Computer Vision and
  Pattern Recognition (CVPR)}, pages 2325--2333, 2016.

\bibitem{Tang2014}
Siyu Tang, Mykhaylo Andriluka, and Bernt Schiele.
\newblock Detection and tracking of occluded people.
\newblock {\em International Journal of Computer Vision}, 110(1):58--69, Oct
  2014.

\bibitem{tian2015deep}
Yonglong Tian, Ping Luo, Xiaogang Wang, and Xiaoou Tang.
\newblock Deep learning strong parts for pedestrian detection.
\newblock In {\em Proceedings of the IEEE International Conference on Computer
  Vision (ICCV)}, pages 1904--1912, 2015.

\bibitem{uijlings2013selective}
Jasper~RR Uijlings, Koen~EA Van De~Sande, Theo Gevers, and Arnold~WM Smeulders.
\newblock Selective search for object recognition.
\newblock {\em International journal of computer vision}, 104(2):154--171,
  2013.

\bibitem{wang2018repulsion}
Xinlong Wang, Tete Xiao, Yuning Jiang, Shuai Shao, Jian Sun, and Chunhua Shen.
\newblock Repulsion loss: Detecting pedestrians in a crowd.
\newblock In {\em Proceedings of the IEEE Conference on Computer Vision and
  Pattern Recognition (CVPR)}, pages 7774--7783, June 2018.

\bibitem{xie2020count}
Jin Xie, Hisham Cholakkal, Rao~Muhammad Anwer, Fahad~Shahbaz Khan, Yanwei Pang,
  Ling Shao, and Mubarak Shah.
\newblock Count-and similarity-aware r-cnn for pedestrian detection.
\newblock In {\em Proceedings of the European Conference on Computer Vision
  (ECCV)}, 2020.

\bibitem{yan2013learn}
Junjie Yan, Zhen Lei, Dong Yi, and Stan Li.
\newblock Learn to combine multiple hypotheses for accurate face alignment.
\newblock In {\em Proceedings of the IEEE International Conference on Computer
  Vision Workshops}, pages 392--396, 2013.

\bibitem{zhang2017citypersons}
Shanshan Zhang, Rodrigo Benenson, and Bernt Schiele.
\newblock Citypersons: A diverse dataset for pedestrian detection.
\newblock In {\em Proceedings of the IEEE Conference on Computer Vision and
  Pattern Recognition (CVPR)}, number~2, page~3, 2017.

\bibitem{zhang2018Occlusionaware}
Shifeng Zhang, Longyin Wen, Xiao Bian, Zhen Lei, and Stan~Z Li.
\newblock Occlusion-aware r-cnn: detecting pedestrians in a crowd.
\newblock In {\em Proceedings of the European Conference on Computer Vision
  (ECCV)}, pages 637--653, 2018.

\bibitem{zhang2018occludedattention}
Shanshan Zhang, Jian Yang, and Bernt Schiele.
\newblock Occluded pedestrian detection through guided attention in cnns.
\newblock In {\em Proceedings of the IEEE Conference on Computer Vision and
  Pattern Recognition (CVPR)}, pages 6995--7003, June 2018.

\bibitem{zhou2019discriminative}
Chunluan Zhou, Ming Yang, and Junsong Yuan.
\newblock Discriminative feature transformation for occluded pedestrian
  detection.
\newblock In {\em Proceedings of the IEEE International Conference on Computer
  Vision (ICCV)}, pages 9557--9566, 2019.

\bibitem{zhou2017multilabelpart}
Chunluan Zhou and Junsong Yuan.
\newblock Multi-label learning of part detectors for heavily occluded
  pedestrian detection.
\newblock In {\em Proceedings of the IEEE International Conference on Computer
  Vision (ICCV)}, pages 3506--3515, Oct 2017.

\bibitem{Zhou_2018_bibox}
Chunluan Zhou and Junsong Yuan.
\newblock Bi-box regression for pedestrian detection and occlusion estimation.
\newblock In {\em Proceedings of the European Conference on Computer Vision
  (ECCV)}, pages 135--151, 2018.

\bibitem{zhou2020NOH-NMS}
Penghao Zhou, Chong Zhou, Pai Peng, Junlong Du, Xing Sun, Xiaowei Guo, and
  Feiyue Huang.
\newblock Noh-nms: Improving pedestrian detection by nearby objects
  hallucination.
\newblock In {\em Proceedings of the 28th ACM International Conference on
  Multimedia}, pages 1967--1975, 2020.

\end{thebibliography}
}

\end{document}